\documentclass[11pt,a4paper]{article}
\usepackage[letterpaper]{geometry}
\usepackage[nohyperref]{emnlp2018}
\usepackage{times}
\usepackage{latexsym}
\usepackage{amssymb}
\usepackage{amsmath}
\usepackage{txfonts}  
\usepackage{tikz}
\usepackage{tikz-qtree}
\usepackage{stmaryrd}
\usepackage{multirow}
\usepackage{graphicx}
\usepackage{url}
\usepackage{mathtools}
\usepackage{comment}

\aclfinalcopy 

\def\WV{W}
\def\MV{\mathbf{M}}

\def\jv{j}
\def\wv{w}
\def\dv{d}
\def\DV{D}
\def\av{a}
\def\bv{b}
\def\GV{\mathbf{G}}
\def\CV{\mathbf{C}} 
\def\cv{\mathbf{c}}
\def\gstar{\GV}
\def\gdstar{\GV_D}
\DeclareMathOperator*{\argmax}{arg\,max}
\def\diag{\mathrm{diag}}

\newcommand{\indicator}[1]{\llbracket #1 \rrbracket}

\newcommand{\condprob}[2]{{\sf P}(#1 \mid #2)}
\newcommand{\modelcondprob}[3]{{\sf P}_{#1}(#2 \mid #3)}

\def\CM{\mathbf{C}}
\def\cv{c}
\def\CV{C}
\def\IV{I}
\def\NV{N}
\def\cv{c}
\def\gl{1} 
\def\gr{2} 
\def\DM{\mathbf{D}}
\def\EM{\mathbf{E}}
\def\VT{\mathbf{V}}
\def\kv{k}
\def\sv{s}

\title{Depth-bounding is effective: Improvements and evaluation of unsupervised PCFG induction}

\author{
Lifeng Jin \\
Department of Linguistics \\
The Ohio State University \\
{\tt jin.544@osu.edu} 
\And
Finale Doshi-Velez \\
Harvard University \\
{\tt finale@seas.harvard.edu} 
\And
Timothy Miller \\
Boston Children's Hospital \& \\
Harvard Medical School \\
{\tt \hspace{-23mm}timothy.miller@childrens.harvard.edu} 
\AND
William Schuler \\
Department of Linguistics \\
The Ohio State University \\
{\tt schuler@ling.osu.edu} 
\And
Lane Schwartz \\
Department of Linguistics \\
University of Illinois at Urbana-Champaign \\
{\tt lanes@illinois.edu} }

\date{}

\begin{document}
\maketitle
\begin{abstract}
There have been several recent attempts to improve the accuracy of grammar induction systems by bounding the recursive complexity of the induction model
\cite{Ponvert2011,Noji2016d,Shain2016,Jin2018}.
Modern depth-bounded grammar inducers have been shown to be more accurate than early unbounded PCFG inducers, but this technique has never been compared against unbounded induction within the same system, in part because most previous depth-bounding models
are built around sequence models, the complexity of which grows exponentially with the maximum allowed depth.
The present work instead applies depth bounds within a chart-based Bayesian PCFG inducer \citep{Johnson}, where bounding can be switched on and off, and then samples trees with and without bounding.%
\footnote{ The public repository can be found at \url{https://github.com/lifengjin/dimi_emnlp18}.
}
Results show that depth-bounding is indeed significantly effective in limiting the search space of the inducer and thereby increasing the accuracy of the resulting parsing model.
Moreover, parsing results on English, Chinese and German show that this bounded model with a new inference technique is able to produce parse trees more accurately than or competitively with state-of-the-art constituency-based grammar induction models.
\end{abstract}

\section{Introduction}

Unsupervised grammar inducers hypothesize hierarchical
structures for strings of words. Using context-free grammars (CFGs) to define these structures, previous attempts at either CFG parameter estimation \cite{Carroll1992,Schabes1992,Johnson} or directly inducing a CFG as well as its probabilities \cite{Liang2009,Tu2012} have not achieved as much success as experiments with other kinds of formalisms \cite{Klein2004a,Seginer2007a,Ponvert2011}. The assumption has been made that the space of grammars is so big that constraints must be applied to the learning process to reduce the burden of the learner \cite{Gold1967,Cramer2007,Liang2009}.

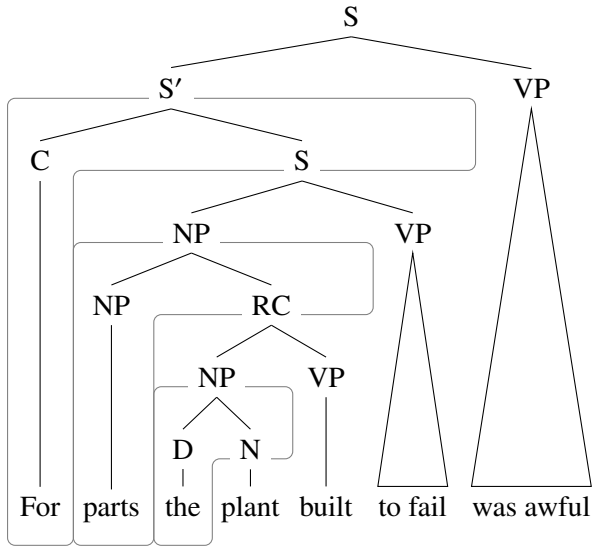
\begin{figure}
\begin{tikzpicture}[baseline,x=2.5em,y=2\baselineskip,level distance=2\baselineskip,sibling distance=3mm]
\tikzstyle{every node}=[anchor=mid,fill=white]
\tikzset{edge from parent path={(\tikzparentnode.south) edge (\tikzchildnode.north)},frontier/.style={distance from root=6.5cm}}
\draw[gray,   rounded corners=1mm] (-4.7,-1) --    (1.7,-1) -- 
                                                   (1.7,-2) --
                                            (-3.8,-2) --
                                            (-3.8,-7.2) --
                                   (-4.7,-7.2) -- cycle;
\draw[gray,   rounded corners=1mm] (-3.8,-3) --    (.3,-3) -- 
                                                   (.3,-4) --
                                            (-2.7,-4) --
                                            (-2.7,-7.2) --
                                   (-3.8,-7.2) -- cycle;
\draw[gray,   rounded corners=1mm] (-2.7,-5) --   (-.8,-5) -- 
                                                  (-.8,-6) --
                                            (-1.9,-6) --
                                            (-1.9,-7.2) --
                                   (-2.7,-7.2) -- cycle;
\Tree[.S [.S$'$ [.C \!\!For\!\! ]
                [.S [.NP [.NP \!\!parts\!\! ]
                         [.RC [.NP [.D \!\!the\!\! ]
                                   [.N \!\!\!plant\!\!\! ] ]
                              [.VP \!\!built\!\! ] ] ]
                    [.VP \edge[roof]; {\!\!to fail\!\!} ] ] ]
         [.VP \edge[roof]; {\!\!was awful\!\!} ] ]
\end{tikzpicture}
\caption{Stack elements after the word {\em the} in a left-corner parse of the sentence {\em For parts the plant built to fail was awful.}}
\label{fig:tree}
\end{figure}

One constraint that has been applied is recursion depth \cite{Schuler2010,Ponvert2011,Shain2016,Noji2016d,Jin2018}, motivated by human cognitive constraints on memory capacity~\cite{chomskymiller63}.
Recursion depth can be defined in a left-corner parsing paradigm~\cite{rosenkratzlewis,johnsonlaird83}.
Left-corner parsers require only minimal stack memory to process left-branching and right-branching structures, but require an extra stack element to process each center embedding in a structure.
For example, a left-corner parser must add a stack element for each of the first three words in the sentence, \textit{For parts the plant built to fail was awful,} shown in Figure~\ref{fig:tree}.
These kinds of depth bounds in sentence processing have been used to explain the relative difficulty of center-embedded sentences compared to more right-branching paraphrases like \textit{It was awful for the plant's parts to fail}.
%

However, depth-bounded grammar induction has never been compared against unbounded induction in the same system, in part because most previous depth-bounding models 
are built around sequence models, the complexity of which grows exponentially with the maximum allowed depth.
In order to compare the effects of depth-bounding more directly, this work extends a chart-based Bayesian PCFG induction model \citep{Johnson} to include depth bounding, 
which allows both bounded and unbounded PCFGs to be induced from unannotated text. 

Experiments reported in this paper confirm that depth-bounding does empirically have the effect of significantly limiting the search space of the inducer. Analyses of this model also show that the posterior samples are indicative of implicit depth limits in the data. This work also shows for the first time that it is possible to induce an accurate unbounded PCFG from raw text with no strong linguistic constraints. With a novel grammar-level marginalization in posterior inference, comparisons of the accuracy of bounded grammar induction using this model against other recent constituency grammar inducers show that this model is able to achieve state-of-the-art or competitive results on datasets in multiple languages.

\section{Related work}

Induction of PCFGs has long been considered a difficult problem \citep{Carroll1992,Johnson,Liang2009,Tu2012}. Lack of success for direct estimation was attributed either to a lack of correlation between the linguistic accuracy and the optimization objective \citep{Johnson}, or the likelihood function or the posterior being filled with weak local optima \citep{Smith2006,Liang2009}. Much of this grammar induction work used strong linguistically motivated constraints or direct linguistic annotation to help the inducer eliminate some local optima. \citet{Schabes1992} use bracketed corpora to provide extra structural information to the inducer. Use of part-of-speech (POS) sequences in place of word strings is popular in the dependency grammar induction literature \citep{Klein2002,Klein2004a,Berg-Kirkpatrick2010,Jiang,Noji2016d}.  Combinatory Categorial Grammar (CCG) induction also relies on POS tags to assign basic categories to words \citep{Bisk2012,Bisk}, among other constraints such as CCG combinators. Other linguistic constraints such as constraints of root nodes \citep{Noji2016d}, attachment rules \citep{Naseem2010} or acoustic cues \citep{Pate2013} have also been used in induction. 

Depth-like constraints have been applied in work by \citet{Seginer2007a} and \citet{Ponvert2011} to help with the search. Both of these systems are successful in inducing phrase structure trees from only words, but only generate unlabeled constituents.

Depth-bounds are directly used by induction models in work by \citet{Noji2016d}, \citet{Shain2016} and \citet{Jin2018}, and are shown to be beneficial to induction.
\newcite{Noji2016d} apply depth-bounding to dependency grammar induction with POS tags. However the constituency parsing evaluation scores they report are low compared to other induction systems. The model in \citet{Shain2016} is a hierarchical sequence model instead of a PCFG. Although depth-bounding limits the search space, the sequence model has more parameters than a PCFG, therefore benefits brought by depth-bounding may be offset by this larger parameter space.

\newcite{Jin2018} also apply depth-bounding to a grammar inducer and induce depth-bounded PCFGs and show that the depth-bounded grammar inducer can learn labeled PCFGs  competitive with state-of-the-art grammar inducers that only produce unlabeled trees. However, because of the cognitively motivated left-corner HMM sampler used in the model, its state space grows exponentially with the maximum depth and polynomially with the number of categories. This renders the transition matrix and the trellis of the inducer too big to be practical in exploring models with higher depth limits, let alone unbounded models. By using Gibbs sampling for PCFGs~\citep{Goodman1998ParsingInside-Out,Johnson}, here described as the inside-sampling algorithm, the state space of the model proposed in this work grows only polynomially with both the maximum depth and the number of categories. 
This allows experiments with more complex models and also achieves a faster processing speed due to an overall smaller state space.

\section{Proposed model}

The model described in this paper follows \newcite{Jin2018} to induce a depth-bounded PCFG by first inducing an unbounded PCFG and then deterministically deriving the parameters of a depth-bounded PCFG from it.
The main difference between this model and the model in \citet{Jin2018} is that they 
use the bounded PCFG to derive
parameters for a 
factored HMM sequence model,
where a forward-filtering backward-sampling algorithm \cite{Carter1996} can be used in inference.
In contrast, the model described in this paper transforms the unbounded PCFG into a bounded PCFG, and then uses the inside-sampling algorithm \cite{Goodman1998ParsingInside-Out} to sample from the posterior of the parse trees given the bounded PCFG in inference.
This section first gives an overview of the model, then briefly reviews the depth-bounding algorithm for PCFGs \cite{VanSchijndel2013,Jin2018}, and finally describes the inference.

As defined in \newcite{Jin2018}, a Chomsky normal form (CNF) unbounded PCFG is a matrix~$\GV$ of binary rule probabilities with one row for each of~$\CV$ parent symbols~$c$ and one column for each of $\CV^2{+}\WV$ combinations of left and right child symbols~$\av$ and~$\bv$, 
which can be pairs of nonterminals or observed words from vocabulary~$\WV$ followed by null symbols $\bot$:
\begin{equation}
\GV = \sum_{\av,\bv,c} \condprob{ c \rightarrow \av \ \bv }{ c } \ \delta_c \ (\delta_\av \otimes \delta_\bv)^\top
\end{equation}
where $\delta_c$ is a Kronecker delta (a vector with value one at index $c$ and zeros elsewhere) and $\otimes$ is a Kronecker product (multiplying two matrices\footnote{or vectors in case $n$ and $p$ equal one} of dimension~$m\times n$ and~$o\times p$ into a matrix of dimension~$mo\times np$ composed of products of all pairs of elements in the operands).
A deterministic depth-bounding transform $\phi$ is then applied to $\gstar$ to create a depth-bounded version $\gdstar$.
A depth-bounded grammar is composed of a set of side- and depth-specific distributions~$\GV_{\sv,\dv}$:
\begin{align}
\GV_\DV = \sum_{\sv \in \{\gl,\gr\}} \sum_{\dv \in \{1..\DV\}} \DM_{\sv,\dv} \, \GV_{\sv,\dv} \, {\EM_{\sv,\dv}}^\top
\end{align}
where side $s \in \{1,2\}$ indicates left (1) or right (2) child.
Categories in $\GV_\DV$ are made to be side- and depth-specific using transforms~$\DM_{\sv,\dv}$ and~$\EM_{\sv,\dv}$:%
\footnote{
Note that this correctly stipulates depth increases for left children of right children.
}
\begin{subequations}
\begin{align}
\DM_{\sv,\dv} &= \delta_\sv \otimes \delta_\dv \otimes \mathbf{I}
\\
\EM_{\gl,\dv} &= \delta_\gl \otimes \delta_{\dv}   \otimes \mathbf{I} \otimes \delta_\gr \otimes \delta_\dv \otimes \mathbf{I}
\\
\EM_{\gr,\dv} &= \delta_\gl \otimes \delta_{\dv+1} \otimes \mathbf{I} \otimes \delta_\gr \otimes \delta_\dv \otimes \mathbf{I}
\end{align}
\end{subequations}

The generative story of this model is as follows.
The model first generates an unbounded grammar $\gstar$ from the Dirichlet prior.
Distributions over expansions~$\condprob{\cv \rightarrow \av \ \bv}{\cv}$ of each category~$\cv$ in this model are drawn from a Dirichlet with symmetric parameter~$\beta$:
\begin{equation}
\GV \sim \mathrm{Dirichlet}( \beta )
\end{equation}
Trees for sentences~$1..\NV$ are each drawn from a PCFG given parameters~$\GV_\DV = \phi(\GV)$:
\begin{equation}
\tau_{1..\NV} \sim \mathrm{PCFG}( \GV_\DV )
\end{equation}
Each tree~$\tau$ is a set~$\{ \tau_\epsilon, \tau_\gl, \tau_\gr, \tau_{\gl\gl}, \tau_{\gl\gr}, \tau_{\gr\gl}, ...\}$ of category labels~$\tau_\eta$ where~$\eta \in \{\gl,\gr\}^*$ is a Gorn address specifying a path of left or right branches from the root.
Categories of every pair of left and right children~$\tau_{\eta \gl}, \tau_{\eta \gr}$ are drawn from a multinomial defined by the grammar~$\GV_\DV$ and the category of the parent~$\tau_\eta$:
\begin{equation}
\tau_{\eta \gl}, \tau_{\eta \gr} \sim \mathrm{Multinomial}( {\delta_{\tau_\eta}\!}^\top \GV_\DV )
\end{equation}
where $\modelcondprob{\GV_\DV}{\av \ \bv}{\wv}=\modelcondprob{\GV_\DV}{\av \ \bv}{\bot}=\indicator{\av,\bv{=}\bot,\bot}$ for~$\wv \in \WV$, and $\indicator{\cdot}$ is an indicator function.

In inference, a Gibbs sampler can be used to iteratively draw samples from the conditional posteriors of the unbounded grammar and the parse trees. For example, at iteration $t$:
\begin{align}
\GV^{t} &\sim \condprob{\GV^{t}}{\tau_{1 .. N}^{t-1}, \sigma_{\tau_{1 .. N}^{t-1}}, \beta}
\\
\tau_{1 \dots N}^{t} &\sim \condprob{\tau^{t}_{1 .. N}}{\GV^{t}_D, \sigma_{\tau_{1 .. N}^t}}
\end{align}
where~$\sigma_\tau$ denotes the terminals in~$\tau$.
These distributions will be defined in Section \ref{sec:gibbs}. 

\subsection{Depth-bounding a PCFG}

This section summarizes the depth-bounding function~$\phi$ for PCFGs described in \citet{VanSchijndel2013} and \citet{Jin2018}. Depth-bounding essentially creates a set of PCFGs with depth- and side-specific categories where no tree that exceeds its depth bound can be generated by the bounded grammar. Because depth increases when a left child of a right child of some parent category performs non-terminal expansion, the probability of such expansions at the maximum depth limit as well as non-depth-increasing expansions beyond the maximum depth limit must be removed from the unbounded grammar. 
\def\hv{\mathbf{h}}
\def\sv{s}
\def\iv{i}
Following \newcite{VanSchijndel2013} and \newcite{Jin2018}, this can be done by iteratively defining a side- and depth-specific containment likelihood~$\hv^{(\iv)}_{\sv,\dv}$
for left- or right-side siblings~$\sv \in \{\gl,\gr\}$
at depth~$\dv \in \{1..\DV\}$
\def\IV{I}
at each iteration~$\iv \in \{1..\IV\}$,
as a vector with one row for each nonterminal or terminal symbol (or null symbol $\bot$) in~$\GV$, containing the probability of each symbol generating a complete yield within depth~$\dv$ as an $\sv$-side sibling:
\begin{subequations}
\begin{align}
\hv^{(0)}_{\sv,\dv}
 &=
  \mathbf{0}
\\
\hv^{(\iv)}_{\gl,\dv}
 &= 
  \begin{cases}
      \GV \, ( \mathbf{1} \otimes \delta_\bot 
                 \ + \ \hv^{(\iv-1)}_{\gl,\dv} \otimes \hv^{(\iv-1)}_{\gr,\dv} )
     & \!\!\text{if } \dv \leq \DV+1
   \\
   \mathbf{0}
     & \!\!\text{if } \dv > \DV+1
  \end{cases}
\\
\hv^{(\iv)}_{\gr,\dv}
 &= 
  \begin{cases}
   \delta_{\mathrm{T}}
     & \!\!\text{if } \dv = 0
   \\
   \GV \, ( \mathbf{1} \otimes \delta_\bot 
              \ + \ \hv^{(\iv-1)}_{\gl,\dv+1} \otimes \hv^{(\iv-1)}_{\gr,\dv} )
     & \!\!\text{if } 0 < \dv \leq \DV
   \\
   \mathbf{0}
     & \!\!\text{if } \dv > \DV
  \end{cases}
\end{align}
\end{subequations}
where `T' is a top-level category label at depth zero.
Following previous work, experiments described in this paper use $I=20$.

A depth-bounded grammar~$\GV_{\sv,\dv}$ can then be defined to be the original grammar~$\GV$ reweighted and renormalized by this containment likelihood:%
\begin{subequations}
\begin{align}
\GV_{\gl,\dv}
 &= \frac{ \GV \, \diag( \mathbf{1} \otimes \delta_\bot
                           \ + \ \hv^{(\IV)}_{\gl,\dv} \otimes \hv^{(\IV)}_{\gr,\dv} ) }
         { \hv^{(\IV)}_{\gl,\dv} }
\\
\GV_{\gr,\dv}
 &= \frac{ \GV \, \diag( \mathbf{1} \otimes \delta_\bot
                           \ + \ \hv^{(\IV)}_{\gl,\dv+1} \otimes \hv^{(\IV)}_{\gr,\dv} ) }
         { \hv^{(\IV)}_{\gr,\dv} }
\end{align}
\end{subequations}

\subsection{Gibbs sampling of unbounded grammars and bounded trees}
\label{sec:gibbs}

As defined above, this model samples iteratively from the conditional posteriors of $\condprob{\gstar}{\tau_{0 .. N},\sigma_{\tau_{0 .. N}}, \beta}$ and $\condprob{\tau_{0 .. N}}{\gdstar, \sigma_{\tau_{0 .. N}}}$ in inference,
extending the Gibbs sampling algorithm for PCFG induction introduced in \citet{Johnson} to depth-bounded grammars.
The below equations will omit the superscript $t$ for the iteration number of inference for clarity.

To sample from the conditional posterior of $\GV$, it is necessary to first sum over all rule applications in all sampled trees:
\begin{align}
\CM_\DV = \sum_{\tau \in \tau_{1..\NV}} \sum_{\tau_{\eta} \in \tau} \delta_{\tau_\eta}\ (\delta_{\tau_{\eta\gl}} \otimes \delta_{\tau_{\eta\gr}})^\top
\end{align}
then remove side- and depth-specificity from category labels:
\begin{align}
\CM =  \sum_{\sv} \sum_{\dv} {\DM_{\sv,\dv}}^\top \CM_\DV \, \EM_{\sv,\dv}
\end{align}
A side- and depth-independent grammar is then sampled from these counts, plus the pseudo-count~$\beta$:
\begin{align}
\GV \sim \mathrm{Dirichlet}(\beta + \CM) 
\end{align}

Inside-sampling \cite{Goodman1998ParsingInside-Out,Johnson} is then used to sample from the posterior of trees $\condprob{\tau_{0 .. N}}{\GV_D, \sigma_{\tau_{0 \dots N}}}$.
Given a depth-bounded grammar and a sentence, this algorithm first constructs the inside chart $\VT \in \mathbb{R}^{L \times L \times C}$, where $L$ is the length of the sentence.
A chart vector 
$\VT_{[\iv,\jv,1..\CV]}$
for the span $i,j$ where $i < j \leq L$ in some sentence $\wv_{1..L}$
is the likelihood $\modelcondprob{\GV_\DV}{w_{i .. j}}{\cv}$ of the span for all side- and depth-specific categories~$\cv$:
\begin{multline}
\VT_{[\iv,\jv,1..\CV]} = 
\\
	\begin{cases}
		\GV_\DV \, (\delta_{\wv_\iv} \otimes \delta_\bot)
          &\text{if}\ \jv{-}\iv=1 \\
        \sum_{\kv} \GV_\DV \, (\VT_{[\iv,\kv,1..\CV]} \otimes \VT_{[\kv,\jv,1..\CV]}) 
          &\text{if}\ 
            \jv{-}\iv > 1 
	\end{cases}
\end{multline}

Trees are sampled iteratively from the top down by first choosing a split point~$\kv_{\iv,\jv}$ for the current span $i,j$ such that $\iv < \kv_{\iv,\jv} < \jv$:
\begin{equation}
\kv_{\iv,\jv} \sim \text{Mul}\left( \sum_\kv \delta_\kv \, \delta_{\cv_{\iv,\jv}}^\top \GV_\DV \, ( \VT_{[\iv,\kv,1..\CV]} \otimes \VT_{[\kv,\jv,1..\CV]} ) \right)
\end{equation}
The algorithm then samples pairs of category labels~$\cv_{\iv,\kv_{\iv,\jv}}$ and~$\cv_{\kv_{\iv,\jv},\jv}$ adjacent at this split point~$\kv_{\iv,\jv}$:
\begin{equation}
\cv_{\iv,\kv}, \cv_{\kv,\jv} \sim \text{Mul}\left( \, \delta_{\cv_{\iv,\jv}}^\top \GV_\DV \, \diag( \VT_{[\iv,\kv,1..\CV]} \otimes \VT_{[\kv,\jv,1..\CV]} ) \, \right)
\end{equation}

Empirically the sampler spends most of its time constructing the inside chart.
%
The model described in this paper therefore
efficiently computes the inside chart using matrix multiplication, which is able to exploit GPU optimization.

\subsubsection*{Efficient inside score calculation}

\begin{figure}
\centering
\includegraphics[width=0.4\textwidth]{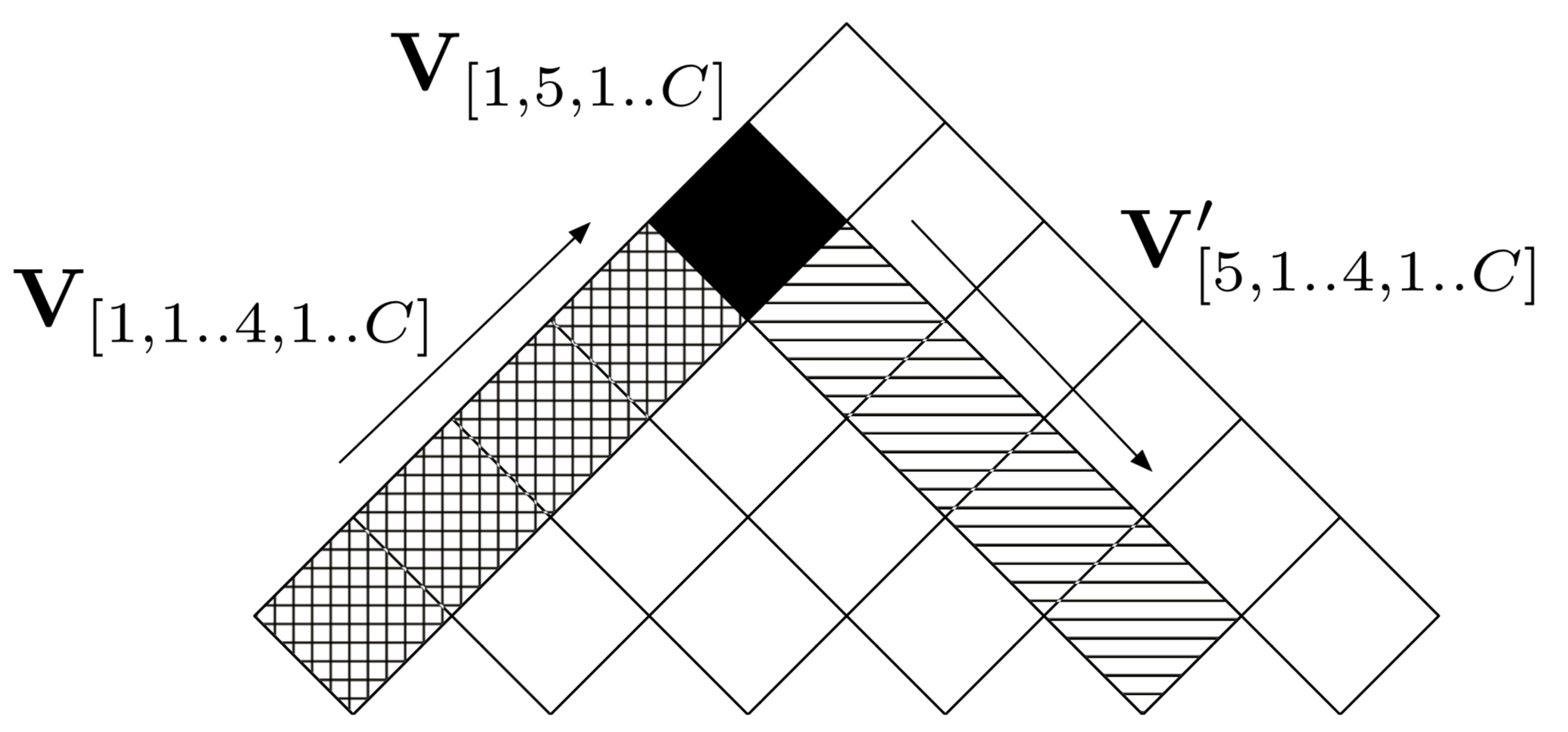}
\caption{Example of matrix multiplication in place of looping over break points for the span (0,5). Each chart cell represents a likelihood vector for the span between $i$ and $j$ where $i$ is the leftmost delimiting index of the span and $j$ the rightmost. The arrows represent the order in which the cells are stored in the chart matrices $\VT$ and $\VT^{\prime}$.}
\label{fig:chart}
\end{figure}

The complexity of the inside algorithm is cubic on the length of the sentence because it has to iterate over all start points $i$, all end points $j$ and all split points $k$ of a span.
For a dense PCFG with a large number of states, the explicit looping is undesirable, especially when it can be formulated as matrix multiplication.
The split point loop is therefore replaced with a matrix multiplication in order to take advantage of highly optimized GPU linear algebra packages like cuBLAS and cuSPARSE, whereas previous work explores how to parse efficiently on GPUs \citep{Johnson11GPU,Canny2013,Hall2014}.

Inside likelihoods are propagated using a copy~$\VT'$ of the inside likelihood tensor~$\VT$ with the first and second indices reversed:
\begin{equation}
\VT'_{[\jv,\iv,\cv]} = \VT_{[\iv,\jv,\cv]}
\end{equation}
This reversal allows the sum over split points~$\kv \in \{\iv{+}1,...,\jv{-}1\}$ to be calculated as a product of contiguous matrices, which can be efficiently implemented on a GPU:
\begin{equation}
\VT_{[\iv,\jv,1..\CV]} = \GV_\DV \, \text{vec}( {\VT_{[\iv,\iv+1..\jv-1,1..\CV]}}^\top \VT'_{[\jv,\iv+1..\jv-1,1..\CV]} )
\end{equation}
where~$\text{vec}(\MV)$ flattens a matrix~$\MV$ into a vector.

\subsection{Posterior inference on constituents}

Prior work \citep{Johnson2007} shows that using EM-like algorithms, which seek to maximize the likelihood of data marginalizing out the latent trees, does not yield good performance. Because trees are the main target for evaluation, it may be preferable to find the most probable tree structures given the marginal posterior of tree structures compared to finding the most probable grammar. Some recent work \citep{Mcclosky2015,Keith2018} explores how to use marginal distributions of tree structures from supervised parsers to create more accurate parse trees. Based on these arguments, this model performs maximum a posteriori (MAP) inference on constituents (PIoC) using approximate conditional posteriors of spans to create final parses for evaluation.

Formally, let $\sigma_{i,j}^\star$ be an MAP unlabeled span of words in a sentence from a corpus $\sigma$,  with start point $i$ and end point $j$, and $\sigma_{i,k}, \sigma_{k,j}$ its possible children. This algorithm iteratively looks for the best pair of children $\sigma_{i,k}^\star, \sigma_{k,j}^\star$ according to the posterior of the children, using all posterior samples. The spans are sentence-specific, but the below equations omit the sentence index for brevity:
\begin{align}
\begin{split}
\sigma_{i,k}^\star, \sigma_{k,j}^\star &= \argmax_{\sigma_{i,k}, \sigma_{k,j}} \condprob{\sigma_{i,k}, \sigma_{k,j}}{\sigma_{i,j}^\star, \sigma} \\
		&= \argmax_{\sigma_{i,k}, \sigma_{k,j}} \int \condprob{\sigma_{i,k}, \sigma_{k,j}, \GV}{\sigma_{i,j}^\star,\sigma} \ d\GV \\
        &\approx \argmax_{\sigma_{i,k}, \sigma_{k,j}} \sum_{\hat{\GV} \sim \condprob{\GV}{\sigma}\!\!\!\!} \condprob{\sigma_{i,k}, \sigma_{k,j}, \hat{\GV}}{\sigma_{i,j}^\star, \sigma}，
\end{split}
\end{align}
where $\sigma$ is the training corpus. Starting from the whole sentence $\sigma_{0, N}$, this algorithm finds the best children for a span from the Monte Carlo estimation of the marginal posterior distribution of children for the span, and then continues to split the found children spans. Because samples from different runs at different iterations can be used to approximate the span posteriors, the process marginalizes out sampled grammars, whole-sentence parse trees and constituent labels to only consider split points for spans. In terms of input and output, the PIoC algorithm takes in posterior samples of trees for a sentence, and outputs an unlabeled binary-branching tree.

There are a few benefits of doing posterior inference on constituents. First, the distribution $\condprob{\sigma_{i,k}, \sigma_{k,j}}{\sigma_{i,j}^\star, \sigma}$ quantifies how much uncertainty there is in splitting a span $\sigma_{i,j}$ at all possible $k$'s. One way of using this uncertainty information is to merge spans where uncertainty is high, effectively weakening or removing the constraint of binary-branching from the grammar inducer. Second, this algorithm produces trees that may not be seen in the samples, potentially helping aggregate evidence across different iterations within a run and across runs. Third, the multimodal nature of the joint posterior of grammars and trees often makes the sampler get stuck at local modes, but doing MAP on constituents may allow information about trees from different modes to come together. If different grammars all consider certain children for a span to be highly likely, then these children should be in the final parse output. Finally, it is a nonparametric way of doing model selection. As will be shown, model selection relies on the log likelihood of the data, but the log likelihood of the data is only weakly correlated with parsing accuracy. Performing PIoC with multiple runs can increase accuracy without depending too heavily on log likelihood for model selection.

\section{Model analysis and evaluation}

The model described above has hyperparameters for maximum depth $\DV$, number of categories $\CV$ and the symmetric Dirichlet prior $\beta$. Following \citet{Jin2018}, this evaluation uses the first half of the WSJ20 corpus as the development set (WSJ20dev) for all experiments. However instead of using the development set only to set the hyperparameters of the model, this evaluation also uses it to explore interactions among parsing accuracy, model fit, depth limit and category domain. The first set of experiments explores various settings of $\DV$ in the hope of acquiring a better picture of how depth-bounding affects the inducer. The second set of experiments uses the value of $\DV$ tuned in the first experiments, and does PIoC on different sets of samples to examine the effect it has on parse quality. Optimal parameter values from these first two experiments are then applied in experiments on English \citep[The Penn Treebank;][]{marcusetal93}, Chinese  \citep[The Chinese Treebank 5.0;][]{Xia2000} and German \citep[NEGRA 2.0;][]{Skut1998} data to show how the model performs compared with competing systems.

Each run in evaluation uses one sample of parse trees from the posterior samples after convergence. Preliminary experiments show that the samples after convergence are very similar within a run and their parsing accuracies differ very little. This evaluation follows \citet{Seginer2007a} by running unlabeled PARSEVAL on parse trees collected from each run. Punctuation is retained in the raw text in induction, and removed in evaluation, also following \citet{Seginer2007a}.

\subsection{Analysis of model behavior}

The first experiment explores the effects of depth-bounding on linguistic grammar quality. The hypothesis is that depth-bounding limits the search space of possible grammars, so the inducer will be less likely to find low-quality local optima where cognitively implausible parse trees are assigned non-zero probabilities, because such local optima would be removed from the posterior by limiting the maximum depth of parse trees to a small number $d$.

\begin{figure}
\centering
\includegraphics[width=0.47\textwidth]{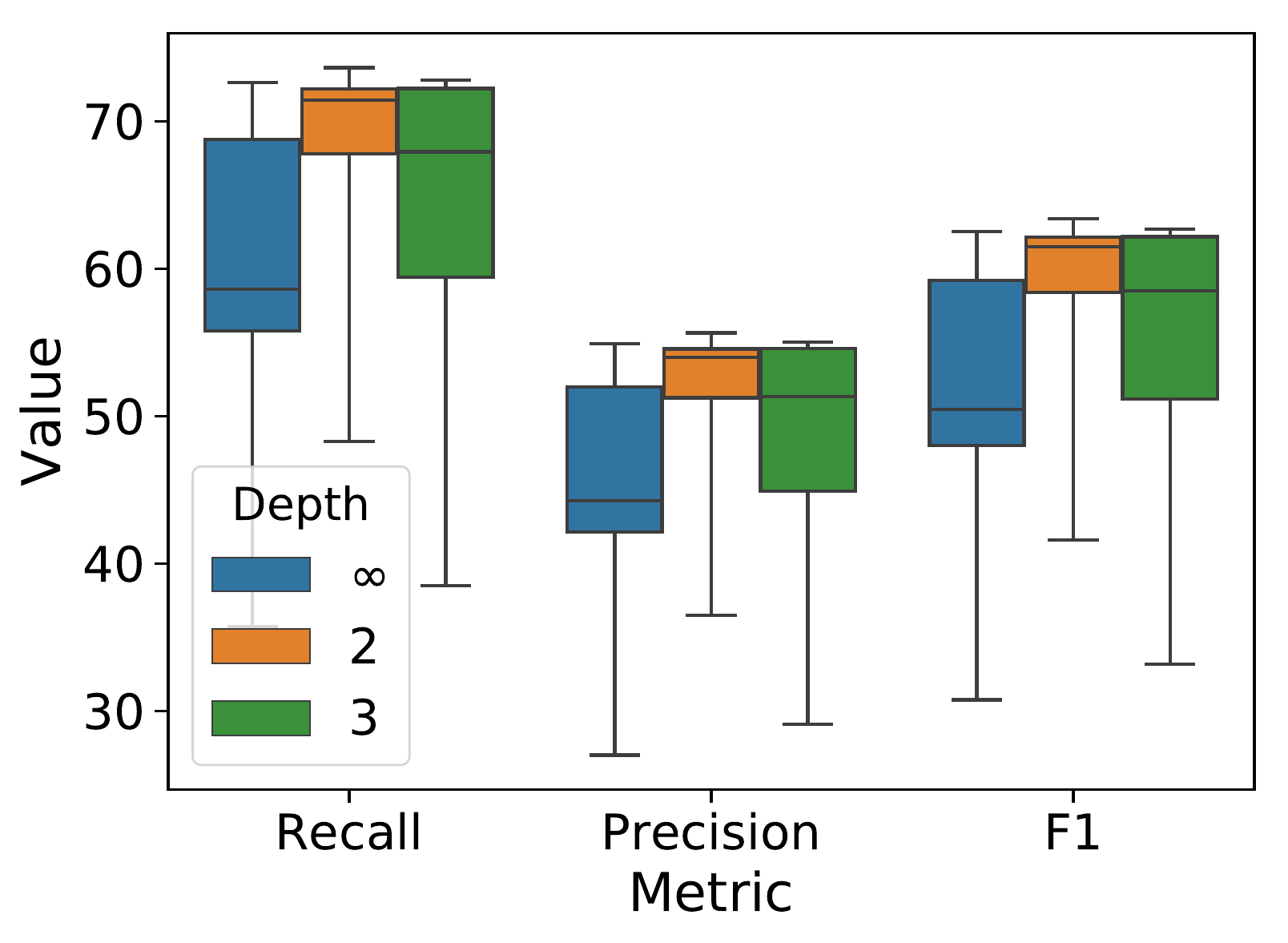}
\caption{PARSEVAL scores for runs with different depth limits. The difference of all PARSEVAL scores between depth $\infty$ and depth 2 is significant ($p$=0.017, Student's $t$ test).}
\label{fig:depth}
\end{figure}

\subsubsection*{The effect of depth-bounding}
Figure \ref{fig:depth} shows the effect of depth bounding using 60 data points of unlabeled PARSEVAL scores from 20 different runs for each of three different depth bounds: 2, 3, and $\infty$ (unbounded). 
The range of possible parsing accuracy scores is very wide, as mapped out by the runs. Although the unbounded model is able to reach the performance upper bound seen from the figure, most of the time its results are in the middle of the range. By bounding the maximum depth to 2, the sampler is able to stay in the region of high parsing accuracy. This may be because the majority of the modes in the region of low parsing accuracy require higher depth limits, and humans who produce the sentences do not have access to those higher depth limits. The difference between depth $\infty$ and depth 2 is significant ($p$=0.017, Student's $t$ test), showing that depth-bounding does have a positive effect on the linguistic grammar quality of the induced grammars. Data from depth 3 also shows a positive trend of inducing better grammars than unbounded.

A purely right-branching baseline achieves an F1 score of 48 on the WSJ20 development dataset.
A majority of induction runs perform better than this baseline,
which indicates that the PCFG induction model with the inside-sampling algorithm is able to find good solutions, most of the time much better than the 
right-branching baseline.
This is especially interesting when the grammar is unbounded with almost no other constraint, which had previously been shown to converge to weak local optima.

\subsubsection*{Correlation of model fit and parsing accuracy}
\begin{figure}
\centering
\includegraphics[width=0.47\textwidth]{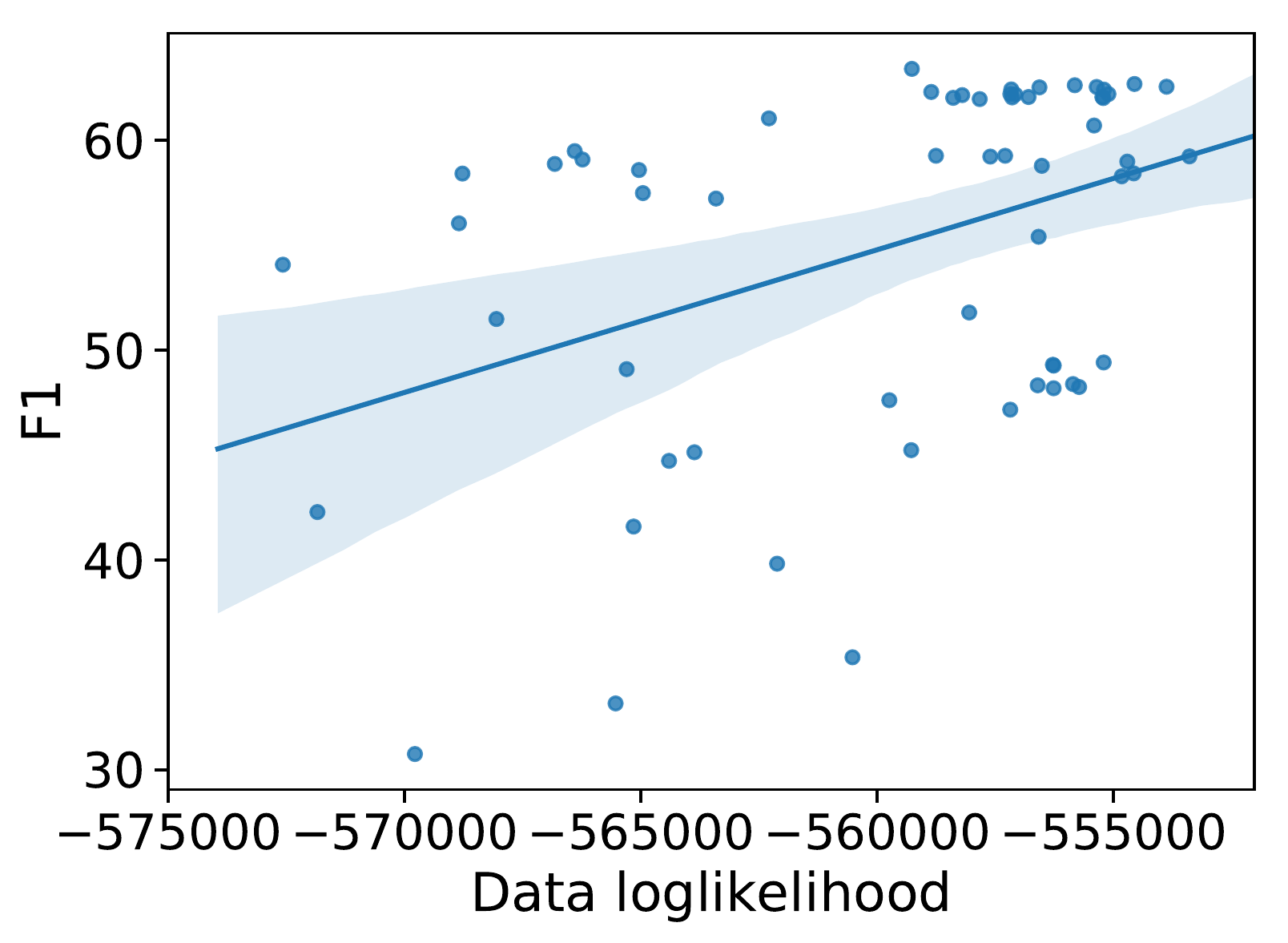}
\caption{The correlation between data likelihood and parsing accuracy of all 60 runs. Calculations show that there is a significant ($p=0.007$) positive correlation (Pearson's $r$=0.39) between data likelihood and parsing accuracy at convergence for our model.}
\label{fig:likelihood}
\end{figure}

Model fit, or data likelihood, has been reported not to be correlated or to be correlated only weakly with parsing accuracy for some unsupervised grammar induction models \citep{Smith2006,Johnson,Liang2009} when the model has converged to a local maximum. Figure \ref{fig:likelihood} shows the correlation between data likelihood and parsing accuracy at convergence for all the runs.  There is a significant ($p=0.007$) positive correlation (Pearson's $r$=0.39) between data likelihood and parsing accuracy at convergence for our model. This indicates that although noisy and unreliable, the data likelihood can be used as a metric to do preliminary model selection.

\subsubsection*{The bounded unbounded PCFG}

\begin{figure}
\centering
\includegraphics[width=0.47\textwidth]{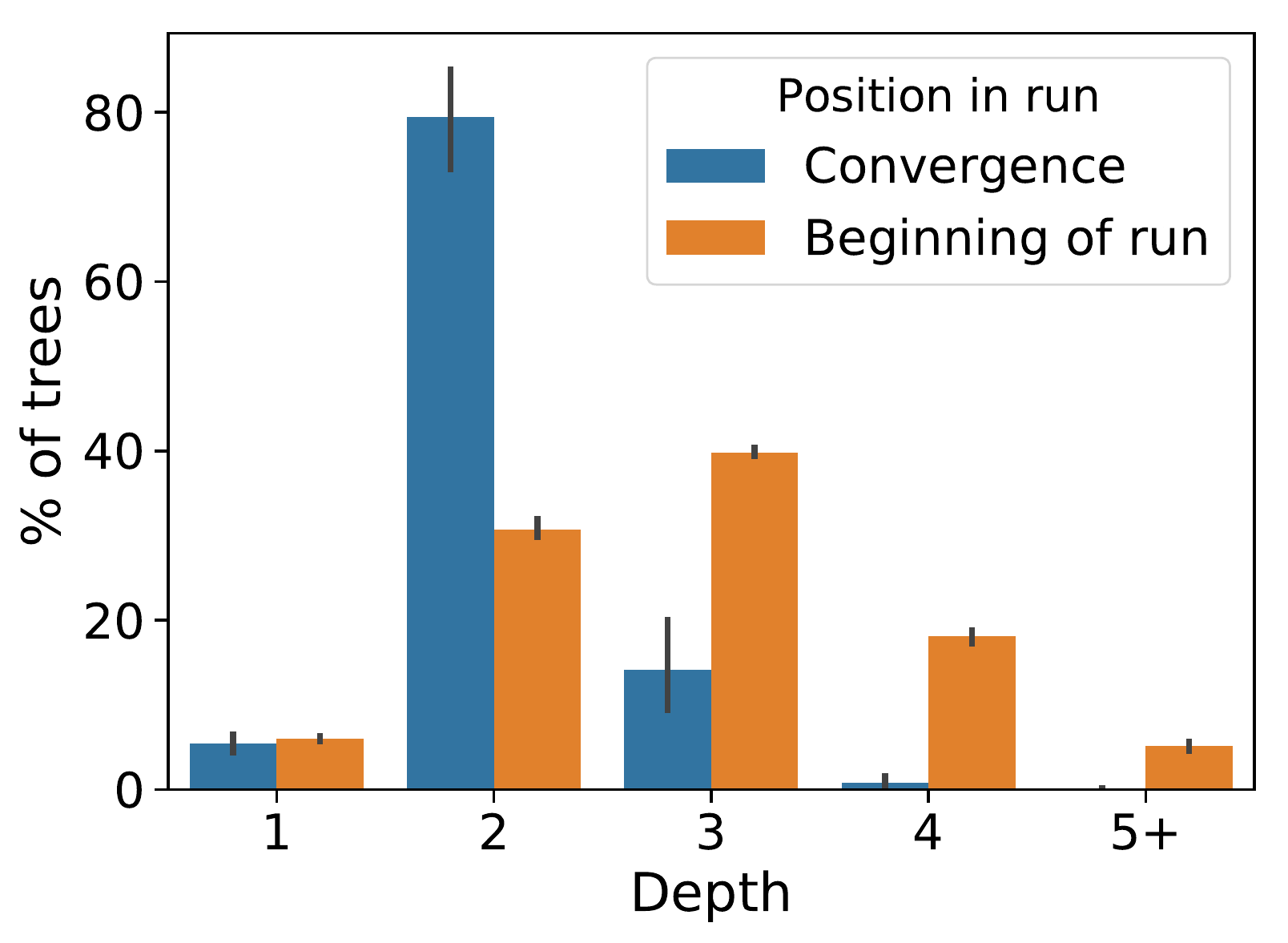}
\caption{The usage of different depths for parse trees in the samples from 20 runs with the unbounded grammar.}
\label{fig:d_dist}
\end{figure}

We also examine the distribution of tree depths in unbounded runs. For a run, we compute the percentage of parse trees with a certain depth, and then examine how these percentages vary across different runs. Theoretically the possible maximum depth of a parse for a sentence is the sentence length divided by 2. For example, a 20-word sentence can have a parse of depth 10 because at least two words are needed to create a new depth with a center embedded phrase, but under most PCFGs this maximally center embedded configuration is not very likely. Figure \ref{fig:d_dist} shows the percentage of tree depths from samples in the beginning of each unbounded run and at convergence. It shows that at the beginning of the sampling process with a random model sampled from the prior, the distribution of parse tree depths seems to be centered around depth 2 and 3, with non-negligible probability mass at other depth levels. At convergence, the distribution of parse tree depths is very peaked with a large portion of the probability mass concentrated at depth 2.
Given that an unbounded PCFG has no constraint on depth, this convergence of the marginal posterior distribution of parse tree depth shows that the depth limit seems to be a natural tendency in the data, rather than an arbitrary preference of corpus annotators. 

\subsection{Posterior uncertainty of constituents}

Experiments were also conducted to determine whether posterior inference on constituents (PIoC) has any effect on parsing accuracy. These experiments use 10 runs on WSJ20dev with depth 2 that have the highest log-likelihoods for exploration.
In this data, some spans have a strikingly higher degree of uncertainty than other spans. 
For example, the posterior probability of splitting the phrase {\em the old story}, into {\em the old} and {\em story} is 0.55, and the probability of splitting it into {\em the} and {\em old story} is 0.45.  Some other spans like {\em use old tools} have virtually no uncertainty in how the inducer evaluates the splits.
Many such spans with high uncertainty are noun phrases, which 
are not annotated with subconstituents in the Penn Treebank annotation.
The parser can therefore avoid precision losses by not splitting constituents with 3 or 4 words if there is large uncertainty in this posterior.%
\footnote{I.e.\ if the difference between the first and the second highest posterior probabilities is smaller than 0.3.}
This experiment only merges spans that would cover 3 or 4 words and leave merging spans with larger coverage to future work.

\begin{table}[t]
\centering
\begin{tabular}{l|c|c|c}
\hline
 System & Rec	&	Prec	&	F1 	\\
\hline
Best	&	73.65	&	55.66	&	63.40 \\
Best w/ PIoC &	73.59	&	56.41	&	63.87	\\
All w/ PIoC &	72.99	&	59.21	&	65.38	\\
All w/ PIoC w/o best &	73.00	&	59.06	&	65.29	\\
\hline
\end{tabular}
\caption{Development results for different systems using posterior inference on constituents (PIoC). }
\label{tab:post}
\end{table}

\begin{table*}[t]
\centering
\begin{tabular}{c|c|c|c|c|c|c|c|c|c}
\hline
\multirow{2}{*}{System}	& \multicolumn{3}{c|}{WSJ20test} & \multicolumn{3}{c|}{CTB20}	& \multicolumn{3}{c}{NEGRA20}	\\
\cline{2-10}
 & Rec	&	Prec	&	F1 & Rec	&	Prec	&	F1 &	Rec	&	Prec	&	F1	\\
\hline
CCL			&	61.7	&	\bf 60.1	&	60.9	&	35.3	&	39.2	& 37.1 &	44.4 & 27.2	& 33.7 \\
UPPARSE			&	40.5	&	47.8	&	43.9	&	33.8	& \bf	44.0 & 38.2 &  55.5	& \bf 41.9	& \bf 47.7 \\
DB-PCFG	&	70.5	&	53.0	&	60.5	&	-	&	-	&	-	&	-&	-& -\\
\hline
this work	&	\bf 73.1	&	55.6	& \bf	63.1	& \bf 43.8	&	35.1 & \bf	38.9 &	\bf 59.1	&	31.2	& 40.8 \\
\hline
\end{tabular}
\caption{PARSEVAL scores for different constituency grammar induction systems. }
\label{tab:multi}
\end{table*}

Table \ref{tab:post} shows parsing results on the WSJ20dev dataset. The {\em Best} result is from an arbitrary sample at convergence of the oracle best run. The {\em Best with PIoC} is the same run, but with PIoC to aggregate 100 posterior samples at convergence. {\em All with PIoC} uses 100 posterior samples from all of the 10 chosen runs, and finally {\em All with PIoC without best} excludes the best run in PIoC calculation. 

There is almost a point of gain in precision going from {\em Best} to {\em Best with PIoC} with virtually no recall loss, showing that the posterior uncertainty is helpful in flattening binary trees. As more samples from the posterior are collected, as shown in {\em All with PIoC without best}, the precision gain is even more substantial. This shows that with PIoC there is no need to  know which sample from which run is the best. Model selection in this case is only needed to weed out the runs with very low likelihood.

\subsection{Multilingual PARSEVAL}

A final set of experiments compare the proposed model with several state-of-the-art constituency grammar induction systems on three different languages. The competing systems are CCL~\citep{Seginer2007a}%
\footnote{\url{https://github.com/DrDub/cclparser}}
and UPPARSE~\citep{Ponvert2011}.%
\footnote{\url{https://github.com/eponvert/upparse}}
We also include the published results of DB-PCFG \citep{Jin2018} on English for comparison.%
\footnote{
We are not able to run DB-PCFG on the other languages due to its substantial resource requirements.
}
The corpora used are the WSJ20test dataset used in \citet{Jin2018}, the CTB20 (sentences with 20 words or fewer from the Chinese Treebank) and NEGRA20 (sentences with 20 words or fewer from the German NEGRA Treebank) datasets used in \citet{Seginer2007a}. All systems are trained and evaluated on the same datasets to ensure fair and direct comparison. Five different induction runs were run on each dataset with the same hyperparameters $D{=}2, \CV{=}15, \beta{=}0.2$ as tuned on the development set, and three runs with the highest likelihood at convergence were chosen for comparison with other models. Parse trees were then calculated using PIoC as previously described, removing punctuation to calculate the unlabeled PARSEVAL scores with EVALB. Multiple runs of CCL and UPPARSE on the same data yield the same results.

Table \ref{tab:multi} shows the unlabeled PARSEVAL scores for the competing systems. The model described in this paper shows strong performance in all languages. On English and Chinese, this model achieves the new state-of-the-art recall and F1 numbers. On German, this model also achieves the best recall scores among all models, showing that more constituents found in the gold annotation are discovered. It is worth noting that the CCL and UPPARSE models do take advantage of additional linguistic constraints, e.g.\ using punctuation as delimiters of constituents. Experiments described in this paper show that this system can perform better than or competitive with these existing models without similar heuristics and constraints.

The model described in this paper performs relatively poorly on precision due to the fact that trees produced by this system are mostly binary-branching with some constituents flattened by PIoC. This issue is most evident on Negra, where fully binary-branching trees have 
nearly twice as many constituents as are annotated in gold.
This puts any system that produces binary-branching trees under a precision celling of 0.51, and F1 celling of 0.675.

\section{Conclusion}
Experiments in this work confirm that depth-bounding does empirically have the effect of limiting the search space of an unsupervised PCFG inducer. Analysis of a depth-bounded model demonstrates desirable engineering properties, including a significant correlation between parsing accuracy and data likelihood, and interesting linguistic properties such as implicit bounding for unbounded grammars. This paper also introduces the Posterior Inference on Constituents technique for model selection and shows for the first time that it is possible to accurately induce a PCFG with no strong universal linguistic constraints.
Comparisons of the proposed model with other state-of-the-art constituency grammar inducers show that this model is able to achieve state-of-the-art or competitive results on datasets in multiple languages.


\section*{Acknowledgments}

The authors would like to thank the anonymous reviewers for their helpful comments.
Computations for this project were partly run 
on the Ohio Supercomputer Center
\shortcite{OhioSupercomputerCenter1987}.
This research was funded by the Defense Advanced Research Projects Agency award HR0011-15-2-0022.
The content of the information does not necessarily reflect the position or the policy of the Government, and no official endorsement should be inferred.

\bibliographystyle{acl_natbib_nourl}
\bibliography{Mendeley_tacl2017}

\end{document}